\documentstyle{llncs}
\begin{document}
\setcounter{page}{320}

\title{LEXIKONEINTR\"AGE F\"UR DEUTSCHE ADVERBIEN}

\author{Ralf \/ Steinberger}
\institute{University of Manchester - Institute of Science and Technology (UMIST)
\thanks{Ich danke dem {\em Kyushu Institute of Technology} (KIT) f\"ur seine 
Einladung zu einem Aufenthalt in Iizuka (Japan). Ein Teil dieses
Beitrags wurde am KIT verfa\ss\/t.} \\ Centre for Computational
Linguistics, P.O. Box 88, Manchester M60 1QD, UK \\
ralf@ccl.umist.ac.uk}

\maketitle

\begin{abstract}

Modifiers in general, and adverbs in particular, are neglected
categories in linguistics, and consequently, their treatment in Natural
Language Processing poses problems. In this article, we present the
dictionary information for German adverbs which is necessary to deal
with word order, degree modifier scope and other problems in NLP. We
also give evidence for the claim that a classification according to
position classes differs from any semantic classification. \\

Angaben im allgemeinen und Adverbien im speziellen sind Stiefkinder der
Linguistik. Auch ihre Behandlung bei der Verarbeitung nat\"urlicher
Sprache bereitet immer wieder Schwierigkeiten. Dieser Bei\-trag zeigt,
welche Information das Lexikon bereitstellen mu\ss\/, um unter anderem
ihre Stellung im Satz zu bestimmen und ihren Skopus erkennen zu
k\"on\-nen. Weiterhin wird gezeigt, da\ss\/ die Stellung von Adverbien
unabh\"angig von deren semantischer Klassifizierung behandelt werden
mu\ss\/. 

\end{abstract}

\section{\"Uberblick}

Conlon \& Evens \cite{Conl} zeigen in {\em Can Computers Handle
Adverbs?}, da\ss\/ die schwie\-rige Wortklasse {\em Adverb} in Natural
Language Processing (NLP) geb\"andigt werden kann, wenn die
Lexikon\-eintr\"age die n\"otige Information enthalten. Sie schlugen
f\"ur das Englische eine semantische und funktionelle Klassifizierung
vor, aus der auch auf das Stellungsverhalten der Adverbien geschlossen
werden kann. Andererseits zeigen Baker \cite{Bake} f\"ur das
Englische und Hoberg \cite{Hobe} f\"ur das Deutsche, da\ss\/ 
Stellungs\-klassen von Adverbien oft nicht mit deren semantischer
Klassifizierung \"ubereinstimmen. Weiterhin kann man f\"ur das
Deutsche wegen seiner freien Wortstellung nicht wie im
Englischen von nur vier Stellungspositionen ausgehen. 

Dieser Beitrag stellt dreizehn lexikalische Merkmale und deren
m\"ogliche Wer\-te vor, die f\"ur die Behandlung von deutschen
Adverbien in NLP notwendig und ausreichend sind. Der erfolgreiche
Einsatz derartiger lexikalischer Eintr\"age im Maschinellen
\"Uber\-setzungs-System CAT2 \cite{Shar} hat gezeigt, da\ss\/ 
Adverbstellung, Grad\-partikel\-skopus\-behandlung, sowie
Homonymdisambiguierung damit zufrie\-den\-stellend gel\"ost werden
k\"onnen.

Wir wollen zuerst den stark uneinheitlich gebrauchten Begriff {\em
Adverb} defi\-nie\-ren (Abschnitt 2), um daraufhin die lexikalischen
Merkmale zu erl\"autern (3) und kurz auf den Formalismus einzugehen,
in dem sie eingesetzt werden (4). Der f\"unfte Abschnitt ist den
Zusammenh\"angen gewidmet, die sich zwi\-schen den Stellungs\-klas\-sen
und den anderen Merkmalen herstellen lassen. Im abschlie\ss\/enden
Ausblick (6) formulieren wir die Vermutung, da\ss\/ die f\"ur
Adverbien erarbeitete Klassifizierung auch auf adverbiale
Pr\"apositionalphrasen (PP) etc.  anwendbar ist.

Eine ausf\"uhrliche Darstellung deutscher Wortstellung im allgemeinen,
und der Behandlung von Adverbien im speziellen, sowie eine Auflistung
von 400 lexikalischen Adverbeintr\"agen, ist in Steinberger
\cite{RS94} enthalten.

\section{Definition des Begriffs `Adverb'} 

In der Vergangenheit wurden f\"ur die Definition der Wortklasse {\em
Adverb} wenigstens 18 verschiedene Kriterien herangezogen \cite{RS94}.
Wir wollen Adverbien hier deshalb folgenderma\ss\/en definieren: 

{\bf Definition:}\/ Der Terminus {\em Adverb} bezeichnet eine
nicht-flek\-tier\-bare Wort\-klasse. Die Nicht\-flektier\-barkeit
unterscheidet das Adverb von den flektierbaren Klassen Verb, Nomen,
Adjektiv, Artikel und Pronomen.  Weitere nicht-flektier\-bare
Klas\-sen sind Pr\"apositionen, Konjunktionen und Interjektionen. Im
Unterschied zu Pr\"apositionen k\"onnen Adverbien, die einen oder
mehrere F\"alle regieren, auch ohne ihre Erg\"anzung auftreten. 
Konjunktionen sind im Unterschied zu Adverbien keine Satzglieder der
untergeordneten S\"atze, die sie einleiten. Interjektionen m\"ussen
syntaktisch isoliert auftreten, und k\"onnen im Gegensatz zu Adverbien
nicht Teil des Satzes sein. 

Obwohl Abt\"onungspartikeln, Gradpartikeln, Modalw\"orter und
pronominale Adverbien nicht explizit erw\"ahnt sind, sind diese
Unterklassen in der Definition eingeschlossen. Die vielfach
diskutierten {\em Satzadjektive}, auch {\em adverbial gebrauchte
Adjektive} etc. genannt, sind jedoch keine Adverbien, da sie zwar nicht
{\em flektiert}, wohl aber {\em flektierbar} sind (vgl. {\em deutlich}
in 1). Dennoch gilt f\"ur sie die gleiche Klassifizierung, weshalb wir
sie im vorliegenden Beitrag einschlie\ss\/en. 

\begin{footnotesize}
\begin{itemize}
\item[1a] Wolfgang spricht {\em deutlich}.
\item[1b] das {\em deutliche} Sprechen
\end{itemize}
\end{footnotesize}

\section{Merkmale von Adverbeintr\"agen}

Die Merkmale betreffen Wortstellung (B, D, E), Skopuserkennung (C, F,
G, H, I), Valenz (J), Komparation (M) und Homonymdisambiguierung (E,
K, L).

{\bf A) Adverb:} Dieses Feld nennt das Lexem. Aus mnemotechnischen
Gr\"unden kann ein Adverb mit Homomymen in anderen Positionsklassen
(vgl. B) mit einem Kommentar versehen werden (vgl. {\em blo\ss\/} in 2):

\begin{footnotesize}
\begin{itemize}
\item[2a] Ach, k\"ame Tina doch {\em blo\ss\/$_{5}$} heute. 
	(Abt\"onungspartikel: Wunsch, Auf\-for\-de\-rung)
\item[2b] {\em Blo\ss\/$_{38}$} Tina kommt heute.  (Gradpartikel: nur, allein) 
\end{itemize}
\end{footnotesize}

{\bf B) Positionsklasse:} Dieses Feld gibt an, in welcher Reihenfolge
die Angaben im Satz aufeinanderfolgen. Adverbiale mit niedrigem Wert
gehen solchen mit h\"oherem Wert voraus. In Anlehnung an Hoberg
\cite{Hobe} unterscheiden wir 44 Stellungskategorien (fortan als Index
zu den Angaben angezeigt), die f\"ur Adverbien ebenso wie f\"ur
adverbial gebrauchte Satzadjektive, NPs und PPs gelten. Sie sind das
Ergebnis einer empirischen \"Uberpr\"ufung von 22 intuitiven Klassen.

Ein Teil der Positionsklassen entspricht der herk\"ommlichen
semantischen Klassifizierung. Ein Gro\ss\/teil ist jedoch scheinbar
von semantischen Inhalten unabh\"angig (vgl. Abschnitt 5). Diese
Nichtentsprechung von Syntax und Semantik erkl\"art, weshalb auf
Semantik basierte Aussagen \"uber die Stellung von Adverbien oftmals
falsch sind. Folgende und weitere Adverbien k\"onnen den
Positionsklassen aufgrund ihrer Semantik zugeteilt werden:

\begin{footnotesize}
\begin{itemize}
\item[19] konditionale Adverbien
\item[22] kausale Adverbien
\item[24] finale Adverbien
\item[27] lokale Adverbien
\item[44] instrumentale Adverbien
\end{itemize}
\end{footnotesize}

Bei temporalen Adverbien m\"ussen sechs Untergruppen unterschieden
werden:

\begin{footnotesize}
\begin{itemize}
\item[26] Zeitraum, -intervall, -erstreckung
\item[33] evaluative temporale Angaben, Zeitpunkt; diese gehen der Klasse 36 voraus, wohin\-gegen die Gruppe 40 der Gruppe 36 folgt.
\item[36] Wiederholung eines Vorgangs; sie gehen 37 voraus.
\item[37] H\"aufigkeit; diese Gruppe folgt der Gruppe 36.
\item[39] temporal-pragmatische Angaben; sie treten oft adjungiert auf.
\item[40] Zeitpunkt oder Zeitraum; diese folgen der Klasse 37.
\end{itemize}
\end{footnotesize}

{\bf C) Angabegro\ss\/gruppen:} Da die 44 Klassen intuitiv
unzug\"anglich sind, und wir auch zur Behandlung des
Gradpartikelskopus Gro\ss\/gruppen brauchen, unterscheiden wir
weiterhin nach pragmatischen (auch existimatorischen), situativen und
{\em verbbezogenen} Angaben ({\em manner}). Die drei Gruppen
entsprechen in etwa der herk\"ommlichen Einteilung (vgl. z.B.
\cite{Enge88}).

Zu den verbbezogenen Angaben geh\"oren Adverbien wie {\em gern$_{43}$}
und Satzadjektive (Gruppe 43), komitative Angaben wie {\em
miteinander$_{42}$}, sowie instrumentale Angaben wie {\em mit dem
Hammer$_{44}$} und {\em damit$_{44}$}. Situative Angaben entsprechen
in etwa den Positionsklassen 19 bis 40. Sie sind eher satz- als
verbbezogen. Zu ihnen geh\"oren alle temporalen Untergruppen, sowie
lokale, kausale, konsekutive, finale etc. Adverbiale. Pragmatische
Angaben entsprechen in etwa den Positions\-klassen 1 bis 18.  Sie sind
im Gegensatz zu den anderen eher sprecherbezogen und dr\"ucken oft ein
Urteil aus. Zu dieser Gruppe geh\"oren viele Abt\"onungspartikeln
(ja$_{1}$, eben$_{5}$, n\"amlich$_{5}$) und Adverbien wie {\em
leider$_{13}$}, {\em ohnehin$_{9}$} und {\em dummerweise$_{14}$}.

Die Einteilung in Gro\ss\/gruppen ist wegen der Entsprechung mit den
Positionsklassen redundant. Auch \"uberlappen die drei Bereiche und sind
deshalb ungenau. Adverbien wie {\em oft$_{37}$} beispielsweise geben nicht nur
eine zeitliche Information (situativ), sondern sagen auch etwas \"uber
die Sprechererwartung aus (existimatorisch). Diese Ann\"aherung ist aber
f\"ur die Erkennung von Skopus (siehe F) n\"utzlich und f\"ur unsere
Zwecke ausreichend.

{\bf D) Rhemaf\"ahigkeit:} Das Merkmal {\em Positionsklasse} (B) dient
der Bestimmung der Abfolge von Adverbien untereinander. Ein weiteres
und ebenso schwie\-riges Problem ist die Entscheidung, wo Adverbien
relativ zu anderen Satzgliedern stehen sollen. Zu dieser Frage gibt es
allerdings keine kurze Antwort, denn die Nat\"urlichkeit von
Wortstellung wird durch ein komplexes System interagierender Faktoren
bestimmt \cite{RS94}. In \cite{RSCo} schlagen wir eine Methode vor,
automatisch S\"atze mit nat\"urlicher Wortstellung zu erzeugen, die
dem Kontext Rechnung tr\"agt, indem sie das spezielle
Stellungsverhalten von thematischen und rhematischen Elementen
ber\"ucksichtigt.

Unter anderem deshalb sollte bei der Satzanalyse festgestellt werden,
welche Elemente rhematisch sind, wozu das Merkmal {\em
Rhemaf\"ahigkeit} ben\"otigt wird. Wir bezeichnen ein Adverb dann als
rhemaf\"ahig, wenn es grunds\"atzlich den Satzfokus tragen kann, das
hei\ss\/t, wenn es dasjenige Element sein kann, das im Satz am
st\"arksten betont ist. {\em Gestern$_{26}$} ist beispielsweise
rhemaf\"ahig (3), wohingegen {\em bereits$_{39}$} nicht fokussierbar
ist. Aus diesem Grund mu\ss\/ in (4) anstatt des Adverbs das Verb
fokussiert werden (Gro\ss\/buchstaben bedeuten Fokus):

\begin{footnotesize}
\begin{itemize}
\item[3] \hspace{1.7mm} 	Professor Nomura sang GEStern$_{26}$.
\vspace{1mm}
\item[4a]  * 			Professor Nomura sang BeREITS$_{39}$.
\item[4b] \hspace{1.7mm} 	Professor Nomura SANG bereits$_{39}$.
\end{itemize}
\end{footnotesize}

Ob ein Adverb in einem bestimmten Fall tats\"achlich den Fokus
tr\"agt, mu\ss\/ bei der Satzanalyse entschieden werden. Hierf\"ur
wurden in letzter Zeit mehrere Vorschl\"age unterbreitet (\cite{Kerp}
\cite{Haji} \cite{RSCo}). 

Die Information, ob ein Adverb grunds\"atzlich rhematauglich ist, kann
nicht aus anderen Merkmalen gefolgert werden, sondern mu\ss\/
Bestandteil des Lexikon\-ein\-trages sein (vgl. Abschnitt 5). Die
Entscheidung \"uber den Merkmalswert mu\ss\/ notwendigerweise intuitiv
entschieden werden, da diese Information nicht objektivierbar ist und
nicht aus einem Korpus extrahiert werden kann. 

{\bf E) Vorfeldf\"ahigkeit:} Manche Adverbien k\"onnen dem finiten
Verb in Verb-Zweit-S\"atzen vorausgehen (+), andere nicht (--). Das
Merkmal wird bei der Synthese deutscher S\"atze ben\"otigt und kann
auch zur Disambiguierung von Homonymen beitragen:

\begin{footnotesize}
\begin{itemize}
\item[5a] \hspace{1.7mm} Einfach$_{43}$ geht das nicht! (adverbial gebrauchtes Satzadjektiv)
\item[5b] \hspace{1.7mm} Er ging einfach$_{18}$ nicht. (pragmatische Angabe)
\item[5c] * 		 Einfach$_{18}$ ging er nicht.
\end{itemize}
\end{footnotesize}

{\bf F) Skopus:} Dieses Merkmal ist in Verbindung mit den drei
folgenden Merkmalen n\"otig, um den Skopus (Bezug) von Gradpartikeln
richtig zu erkennen und wieder\-zu\-geben. Skopus liegt vor, wenn die
Gradpartikel ein anderes Element entweder graduiert, d.h. auf eine
Skala abbildet, oder fokussiert. Wenn ein solcher Bezug vorliegt,
bilden Gradpartikel und graduiertes Element, mit wenigen Ausnahmen
(siehe H), eine Konstituente, die alleine im Vorfeld stehen kann.

Gradpartikeln unterscheiden sich darin, auf welche Kategorien sie sich
bezie\-hen k\"onnen. M\"ogliche Bezugskonstituenten sind:

\vspace{2mm}
\begin{footnotesize}
\begin{tabular}{l l}
{\bf s: }    & 	Satz \\
{\bf ap: }    &	Adjektivphrase \\
{\bf npp:} &	NP und PP \\
{\bf cp:} &	Kardinalphrase, z.B. {\em zehn} in {\em rund zehn} \\
{\bf neg:} &	Negation, z.B. {\em nicht} in {\em gar nicht} \\
{\bf conj:} &	Konjunktion, z.B. {\em wenn} in {\em nur wenn} \\
{\bf man:} &	verbbezogene Angabe ({\em manner}), z.B. {\em gern$_{43}$} in {\em sehr gern} \\
{\bf sit:} &	situative Angabe, z.B. {\em gestern$_{26}$} und {\em deshalb$_{22}$}, in {\em nur gestern/deshalb} \\
{\bf pragm:} & pragmatische Angabe, z.B. {\em wahrscheinlich$_{12}$} in {\em sehr/ganz wahrscheinlich} \\
\end{tabular}
\end{footnotesize}
\vspace{3mm}

Das Skopusmerkmal ist n\"otig, da nicht alle Gradpartikeln sich auf
alle Elemente beziehen k\"onnen. {\em Rund$_{16}$} beispielsweise kann
sich nur auf Kardinalzahlen, nicht aber auf NPs (*~rund die M\"anner),
Adjektive, S\"atze etc. beziehen. {\em Nur$_{38}$} andererseits kann
sich auf alle drei Angabearten (nur ungern$_{43}$, nur heute$_{26}$,
nur teilweise$_{11}$), die Kategorie npp (nur der Mann, nur f\"ur den
Fall), Kardinalzahlen (nur zwei) und Konjunktionen (nur wenn) sowie
den ganzen Satz (er schmollt nur) beziehen. Allen Angaben, die keinen
spezifischen Bezug haben, haben wir den default-Wert ``s'' zugewiesen.

{\bf G) Pre/Post:} Die meisten Gradpartikeln gehen dem modifizierten
Element direkt voraus (pre). Einige wenige k\"onnen ihm jedoch auch
folgen (post) ({\em dage\-gen$_{6}$} in 6) oder beide Positionen
einnehmen (both) wie {\em noch$_{39}$} in (7):

\begin{footnotesize}
\begin{itemize}
\item[6] Alle Leute hassen Boris. Melina {\em dagegen$_{6}$} liebt ihn. 
\vspace{1mm}
\item[7a] Noch$_{39}$ gestern$_{26}$ war Vah\'e hier.
\item[7b] Gestern$_{26}$ noch$_{39}$ war Vah\'e hier.
\end{itemize}
\end{footnotesize}

{\bf H) Distanz:} Einige wenige Gradpartikeln k\"onnen in
Distanzstellung (+) zu dem Element stehen, auf das sie sich beziehen,
das hei\ss\/t, sie k\"onnen von ihrem Bezugselement durch ein oder
mehrere andere W\"orter getrennt sein:

\begin{footnotesize}
\begin{itemize}
\item[8] {\em Drei} Hamburger hat Vah\'e gestern {\em nur$_{38}$} verdr\"uckt. (nur drei)
\end{itemize}
\end{footnotesize}

{\bf I) Graduierbarkeit:} Nicht alle Elemente lassen sich graduieren. 
Diese Tatsache mu\ss\/ durch ein separates Merkmal ausgedr\"uckt
werden. Das situative Adverb {\em heute$_{26}$} beispielsweise
l\"a\ss\/t sich von {\em nur$_{38}$} modifizieren (+) (nur heute). Das
ebenfalls situative {\em nun$_{26}$} ist hingegen nicht modifizierbar
(--) (* nur nun). 

Die Verwendung der beiden Merkmale {\em Skopus} und {\em
Graduierbarkeit} ist nicht ausreichend, um f\"ur jede Kombination von
Angaben bestimmen zu k\"onnen, ob sie sich aufeinander beziehen, da
manche Elemente zwar grunds\"atzlich graduierbar sind, aber nicht mit
jeder Gradpartikel kombiniert werden k\"onnen. {\em Ganz$_{43}$} kann
beispielsweise einige pragmatische Angaben modifizieren (ganz
sicher$_{12}$), es ist aber nicht mit {\em eher$_{6}$} kombinierbar
(*~{\em ganz eher}; vgl. jedoch {\em viel eher}). Wir sind jedoch der
Ansicht, dass die Unterspezifizierung f\"ur unsere Zwecke ausreicht,
da Kombinationen wie {\em ganz eher} als Eingabe f\"ur Maschinelle
\"Ubersetzungs-Systeme etc. so gut wie ausgeschlossen werden k\"onnen.
Das Zusammenspiel zwischen Skokus (F) und Graduierbarkeit wird in
Abschnitt 4 noch einmal aufgegriffen.

{\bf J) Valenz:} Unsere Definition von Adverbien in 2 l\"a\ss\/t deren
Valenz zu. Das Merkmal hat die m\"oglichen Werte Genitiv, Dativ und
Akkusativ. {\em Abseits$_{27}$} beispielsweise kann mit Genitiv
auftreten:

\begin{footnotesize}
\begin{itemize}
\item[9a] Assunta wohnt abseits$_{27}$.
\item[9b] Assunta wohnt abseits$_{27}$ des Dorfes.
\end{itemize}
\end{footnotesize}
\vspace{1mm}

{\bf K) Pr\"adikativer Gebrauch:} Dieses Merkmal dr\"uckt aus, ob eine
Angabe pr\"a\-di\-ka\-tiv gebraucht werden kann (+) oder nicht (--). 
Mehrere homonyme An\-ga\-ben k\"onnen hierdurch disambiguiert werden, wenn
eine von ihnen bez\"uglich dieses Merkmals einen positiven und die
andere einen negativen Wert hat (so$_{22/43}$, gleich$_{18/33}$,
etc.):

\begin{footnotesize}
\begin{itemize}
\item[10a] \hspace{1.7mm} Tina war so$_{43}$. (Proform f\"ur verbbezogene Angaben wie {\em neugierig$_{43}$})
\item[10b] * Tina war so$_{22}$. (Grund: {\em somit})
\end{itemize}
\end{footnotesize}
\vspace{1mm}

{\bf L) Negierbarkeit:} Dieses Merkmal dr\"uckt aus, ob Angaben auf
die Negationspartikel {\em nicht$_{41}$} folgen k\"onnen (+) oder
nicht (--):

\begin{footnotesize}
\begin{itemize}
\item[11a] \hspace{1.7mm} Wolf kann gerade$_{33}$ nicht$_{41}$ gehen. (temporales {\em gerade}: {\em jetzt})
\item[11b] * Wolf kann nicht$_{41}$ gerade$_{33}$ gehen. 
\vspace{1mm}

\item[12a] \hspace{1.7mm} Wolf kann nicht$_{41}$ gerade$_{43}$ gehen. (verbbezogene Angabe: {\em geradlinig})
\item[12b] * Wolf kann gerade$_{43}$ nicht$_{41}$ gehen.
\end{itemize}
\end{footnotesize}
\vspace{1mm}

{\bf M) Komparierbarkeit:} Nur wenige Adverbien, wie etwa {\em
oft$_{37}$}, sind komparierbar und erhalten somit den positiven Wert
``+''. Soll die NLP-Anwendung die Komparativ- und Superlativformen
auch erkennen und generieren k\"onnen, so ist eine Reihe weiterer
morphologischer Merkmale n\"otig (vgl. \cite{Gaut}).

Tabelle 1 zeigt ein paar exemplarische Eintr\"age.

\begin{table*}
\begin{center}
\caption {Einige exemplarische Adverbeintr\"age}
\vspace{2pt}
\begin{footnotesize}
\begin{tabular}{|c|c|c|c|c|c|c|c|c|c|c|c|c|}
\hline
\hline
A & B & C & D & E & F & G & H & I & J & K & L &M  \\
\hline
\hline
abseits & 27 & sit & + & + & s & - & - & + & Gen & + & + &-  \\
\hline
bereits & 39 & sit & - & - &s,man,sit,ap,npp,cp& both & - & - & - & - & - &-  \\
\hline
blo\ss\/ (nur) & 38 & sit & - & + &s,man,sit,ap,npp,cp& pre & + & - & - & - & + &-\\
\hline
deshalb & 22 & sit & - & + & s & - & - & + & - & - & + &-  \\
\hline
einfach & 18 & pragm & - & - & s & - & - & + & - & - & + &-  \\
\hline
einfach & 43 & man & + & + & s & - & - & + & - & + & + &+  \\
\hline
gestern & 26 & sit & + & + & s & - & - & + & - & + & - &-  \\
\hline
ja (Abt\"onung) & 1 & pragm & - & - & s & - & - & - & - & - & - &-  \\
\hline
leider & 13 & pragm & - & + & s & - & - & - & - & - & - &-  \\
\hline
oft & 37 & sit & + & + & s & - & - & + & - & + & + &+  \\
\hline
rund & 16 & pragm & - & - & cp & pre & - & - & - & - & - &-  \\
\hline
sehr & 43 & man & + & - &s,man,sit,pragm,ap& pre & - & - & - & - & +
&-  \\ [2ex]
\hline
\hline
\end{tabular} 
\end{footnotesize}
\end{center}
\end{table*}

Die meisten dieser Merkmale sind syntaktischer Natur. Die
Unterscheidung zwischen syntaktischen, semantischen und pragmatischen
Merkmalen ist aller\-dings nicht immer durchf\"uhrbar. {\em
Graduierbarkeit} ist beispielsweise eine semantische Eigenschaft, die
sich insofern auf die Syntax auswirkt, als Gradpartikel und graduierte
Phrase eine Konstituente bilden, die im Vorfeld stehen kann. Die
Zugeh\"origkeit des Merkmals {\em Rhemaf\"ahigkeit}, das syntaktische
wie pragmatische Implikationen haben kann, ist ebenfalls nicht
eindeutig.

Es hat sich gezeigt, da\ss\/ diese dreizehn Merkmale f\"ur Maschinelle
\"Ubersetzung ausreichend sind. F\"ur andere Anwendungen k\"onnte es
jedoch n\"otig sein, weiterhin Angaben \"uber die Kompatibilit\"at von
Angaben zu machen, wie von Conlon \& Evens \cite{Conl} vorgeschlagen. Auch
k\"onnen manche Adverbien eine Auswirkung auf Tempus und Aspekt haben
(13 vs. 14). In diesem Fall m\"ussen zus\"atzliche Merkmale herangezogen
werden:

\begin{footnotesize}
\begin{itemize}
\item[13a] \hspace{1.7mm}   Archana liest gerade.
\item[13b] \hspace{1.7mm}   Archana is reading.
\item[13c] *		    Archana reads. (inkorrekt als \"Ubersetzung von 13a)
\vspace{1mm}

\item[14a] \hspace{1.7mm}   Archana liest oft.
\item[14b] \hspace{1.7mm}   Archana often reads.
\item[14c] *		    Archana is often reading.
\end{itemize}
\end{footnotesize}

\section{Einsatz der Merkmale in der Maschinellen \"Ubersetzung}

Lexikalische Eintr\"age mit diesen Merkmalen werden im Maschinellen
\"Uber\-setz\-ungs-System CAT2 \cite{Shar} erfolgreich eingesetzt. CAT2
ist ein transfer- und unifikationsbasiertes multilinguales
\"Uber\-setzungs-System, das am {\em Institut f\"ur Angewandte
Informationsforschung} in Saarbr\"ucken seit 1987 als Seitenlinie von
EUROTRA entwickelt und in mehreren europ\"aischen Projekten eingesetzt
wird (vgl. auch \cite{Stre}). Die Implementierung ist in Steinberger 
\cite{RS92a} (Wortstellung) und \cite{RS92b} (\"Ubersetzung von 
Gradpartikelskopus) detailliert beschrieben. An dieser Stelle soll nur
exemplarisch angedeutet werden, wie Gradpartikelskopuserkennung in CAT2
behandelt werden kann. Die Merkmale sind jedoch als weitgehend
for\-malis\-musunabh\"angig anzusehen, soda\ss\/ eine andere Verwendung
m\"oglich ist.

CAT2 weist einer Sequenz von Lexemen bei der Analyse eine Struktur zu,
wenn die in Merkmalen ausgedr\"uckte lexikalische Information mit den
Phrasenstrukturregeln unifiziert. In unserer Implementierung
\cite{RS92b} beschreibt beispiels\-weise die Regel f\"ur komplexe 
Adverbialphrasen wie {\em sehr oft} die Struktur {\em AdvP
${\Rightarrow}$ AdvP~Adv}. Um die Analyse von Adverbsequenzen wie {\em
des\-halb oft} als Gra\-du\-ierung zu vermeiden und anstatt dessen zwei
unabh\"angige Adverbialphrasen zu bilden, sind in der AdvP-Regel
Bedingungen formuliert: Erstens muss das graduierte Element das
Merkmal besitzen, da\ss\/ es graduierbar ist (I), und zweitens mu\ss\/
die Kategorie (C) des graduierten Elements mit einem der Skopuswerte
(F) des graduierenden Adverbs \"ubereinstimmen (Variablenbindung). 
Sind diese Bedingungen nicht erf\"ullt, so greift die Regel nicht und
der Sequenz wird diese Modifikationsstruktur nicht zugewiesen.

In CAT2 sammelten wir alle Merkmale, die f\"ur die Behandlung von
Gradpartikelskopus gebraucht werden, in einem \"ubergeordneten
Merkmal, so da\ss\/ die gesamte Information in einem Schritt in den
Mutterknoten hochperkuliert werden kann. Diese Anordnung ist jedoch
keineswegs verbindlich. Selbst in Formalismen wie EUROTRA \cite{Bech},
das keine hierarchischen Merkmalsstrukturen erlaubt, ist eine
Implementierung \"ahnlich der unseren in \cite{RS92b} m\"oglich.

\section{Verallgemeinerungen bez\"uglich der Positionsklassen}

Obwohl Hoberg \cite{Hobe} bereits feststellt, da\ss\/ die
Positionsklassen nicht in allen F\"allen mit semanti\-schen Klassen
identisch sind, erwarteten wir, da\ss\/ wenigstens die Elemente
mancher Positions\-klassen in ihren Merkmalswerten
\"uber\-ein\-stim\-men. Insbesondere hatten wir die Intuition, da\ss\/
verbbezogene Angaben immer rhematisch sein k\"onnen, wohingegen
pragmatische Angaben keine m\"oglichen Rhemata sind.  Weiterhin nahmen
wir an, da\ss\/ Elemente, die graduiert werden k\"onnen, auch
negierbar sind, und umgekehrt.

Aus diesem Grund ordneten wir die 400 kodierten Eintr\"age in
Steinberger \cite{RS94} nach Positionsklassen, um zu sehen, ob unsere
Intuition den Tatsachen entsprach. Tabelle 2 zeigt diejenigen
Merkmalswerte, bez\"ug\-lich derer die Positionsklassen homogen sind. 
Es f\"allt auf, da\ss\/ die niedrigen Positionsklassen (1 bis 7), die
so gut wie keine semantischen Gemeinsamkeiten aufweisen, sondern reine
Stellungsklassen sind, fast durchwegs ihre Eigenschaften teilen.

\begin{table*}
\begin{center}
\caption {Verallgemeinerungen bez\"uglich der Positionsklassen}

\vspace{2pt}
\begin{tabular}{|c|c|c|c|c|c|c|c|}
\hline
\hline
Position & Anzahl & Vorfeld & Komp. & Neg. & Grad. & Rhema & Pr\"ad. \\
\hline
B & & E & M & L & I & D & K \\
\hline
\hline
1 & 2 & - & - & - & - & - & - \\
\hline
2 & 4 & - & - & - & - & - & - \\
\hline
3 & 2 & - & - & - & - & - & - \\
\hline
4 & 1 & - & - & - & - & - & - \\
\hline
5 & 7 & - & - & - & - & - & - \\
\hline
6 & 12 & + & - & - & - 1 & - & - \\
\hline
7 & 18 & + & - & - & - & - 1 & - \\
\hline
8 & 10 & & - & - 1 & - 1 & & \\
\hline
9 & 6 & & - & - & - & & - \\
\hline
11 & 11 &  & - & - 1 &  & - 1 & - 1 \\
\hline
12 & 22 & + 1 & - & - &  &  &  \\
\hline
13 & 7 & + & - & - & - & - & - \\
\hline
14 & 13 &  & - &  &  &  &  \\
\hline
15 & 2 &  & - &  & - & + & + \\
\hline
16 & 7 &  & - & - & - & - & - \\
\hline
17 & 4 & - & - & - & - & - & - \\
\hline
18 & 9 & & - & & & - 1 & - \\
\hline
19 & 1 & + & - & - & + & - & - \\
\hline
20 & 2 & + & - & - & - & + & - \\
\hline
21 & 3 & + & - & & + & - & - \\
\hline
22 & 21 & + & - & & & &  \\
\hline
24 & 3 & + & - & + & + & &  \\
\hline
25 & 2 & + & - & + & + & - &  \\
\hline
26 & 46 & + 3 & - 2 & & & &  \\
\hline
26/40 & 18 & + & - 1 & + 1 & + 1 & &  \\
\hline
27 & 43 & + 3 & - & & & &  \\
\hline
28 & 1 & + & - & - & - & - & - \\
\hline
29 & 2 & + & - & + & + & &  \\
\hline
30 & 2 & + & - & + & + & &  \\
\hline
31 & 1 & + & - & + & - & + & + \\
\hline
33 & 18 & - 2 & - & & & &  \\
\hline
34 & 7 & & - & - 1 & & & - \\
\hline
35 & 5 & & - & & & &  \\
\hline
36 & 5 & + & - & & & + &  \\
\hline
37 & 29 & + & & & & &  \\
\hline
38 & 5 & & - & & & - & - \\
\hline
39 & 5 & & - & & & &  \\
\hline
40 & 6 & + & & & & &  \\
\hline
41 & 2 & + & - & + & + & + & - \\
\hline
42 & 1 & + & - & + & + & + & - \\
\hline
43 & 33+ & & & & & &  \\
\hline
44 & 1 & + & - & + & + & - & + \\
\hline
\hline
\end{tabular} 
\end{center}
\end{table*}

Die Werte ``+'' und ``--'' in Tabelle 2 bedeuten, da\ss\/ ausnahmslos
alle Elemente der Klasse einen positiven oder negativen Wert haben. 
Folgt den Zeichen eine Zahl, so deutet diese an, wieviele Ausnahmen es
zu dieser Tendenz gibt. Die Spalte ``Anzahl'' zeigt, wieviele Elemente
den einzelnen Klassen angeh\"oren. Die Positions\-klasse 7
beispielsweise ({\em andererseits, au\ss\/erdem, weiterhin} etc.)
enth\"alt 18 Elemente, von denen alle bis auf eines kein potentielles
Rhema sind. Ist ein Feld leer, so bedeutet das, da\ss\/ die Angabe
bez\"uglich dieses Merkmals heterogen ist. Die Positions\-klasse 26/40
beinhaltet Elemente, die sowohl 26 als auch 40 angeh\"oren. Von der
Angabeklasse 43, die vor allem adverbial gebrauchte Adjektive
beinhaltet und somit offen ist, haben wir nur 33 Eintr\"age
ber\"ucksichtigt. Dies ist durch ein ``+'' angedeutet. Die
Positionsklassen 10, 23 und 32 k\"onnen nur durch PPs realisiert werden.

Aus der \"Ubersicht ist erkennbar, da\ss\/ Verallgemeinerungen
bez\"uglich der Merkmalswerte fast nur f\"ur die Klassen mit wenigen
Mitgliedern gemacht werden k\"onnen. Es sind nur wenige
Allgemeinaussagen bez\"uglich der Adverbien gr\"o\ss\/erer
Positionsklassen m\"oglich, wenngleich es einige Tendenzen gibt. Wir
k\"onnen somit zu Hobergs \cite{Hobe} Feststellung, da\ss\/
Positions\-klassen und semanti\-sche Klassen nicht notwendigerweise
\"uber\-ein\-stimmen, hinzuf\"ugen, da\ss\/ Positions\-klassen auch
bez\"uglich der hier er\-w\"ahn\-ten Merkmale meist heterogen sind. 
F\"ur NLP-Lexikographen bedeutet das, da\ss\/ Adverbeintr\"age
individuell erstellt werden m\"ussen. Andererseits d\"urfte die Anzahl
der Adverbien trotz einiger produktiver Derivationsmittel beschr\"ankt
sein, so da\ss\/ sich der Kodieraufwand in Grenzen h\"alt.

\section{Ausblick}

Unsere Untersuchung beschr\"ankte sich aus praktischen Gr\"unden auf
Adverbien und {\em Satzadjektive}. Wir vermuten jedoch, da\ss\/ diese
Merkmale auch f\"ur NP- und PP-Adverbiale relevant sind. Wir hoffen
weiterhin, da\ss\/ die Klassifikation zur Analyse dieser Angabegruppen
beitragen kann, da sie ein Klassifikationsraster in Form der
relevanten Merkmale und ihrer m\"oglichen Werte bietet. Es w\"are nun
etwa zu pr\"ufen, wie die Pr\"apositionen auf die Positionsklassen
verteilt sind, und ob semantische Merkmale der Nomen f\"ur deren
Verteilung auf die Positionsklassen relevant sind. Eine andere Frage,
die sich aufdr\"angt ist die, ob alle Adverbien, die mithilfe
derselben Derivationsmittel (-seits, -hal\-ber, -weise, etc.) gebildet
wurden, ihre Merkmale mit den anderen Adverbien derselben Klasse
teilen.


\end{document}